\documentclass[journal]{IEEEtran}
\ifCLASSINFOpdf
\else
\fi
\usepackage{url}

\usepackage{epsfig}
\usepackage{graphicx}
\usepackage{amsmath}
\usepackage{amssymb}
\DeclareMathOperator*{\argmax}{argmax}

\usepackage{algorithm}
\usepackage{algorithmic}

\usepackage{multirow}
\usepackage{epstopdf}

\usepackage{color}

\begin{document}
%
\title{A General Model for Robust Tensor Factorization with Unknown Noise}
%
%
%

\author{ Xi'ai Chen,~\IEEEmembership{Student Member,~IEEE,}
         Zhi Han,~\IEEEmembership{Member,~IEEE,}
         Yao Wang,~
         Qian Zhao,~
        Deyu Meng,~\IEEEmembership{Member,~IEEE,}
        Lin Lin
        and Yandong Tang,~\IEEEmembership{Member,~IEEE}
\thanks{Xi'ai Chen, Zhi Han and Yandong Tang are with State Key Laboratory of Robotics,
 Shenyang Institute of Automation, Chinese Academy of Sciences, Shenyang, LiaoNing, 110016 China
 e-mails: chenxiai@sia.cn, hanzhi@sia.cn, ytang@sia.cn.}
\thanks{Xi'ai Chen is with University of Chinese Academy of Sciences, Beijing, 100049 China
 e-mail: chenxiai@sia.cn.}
\thanks{Yao Wang, Qian Zhao, Lin Lin and Deyu Meng are with School of Mathematics and Statistics, Xi'an Jiaotong University, Xi'an, Shaanxi 710049 China. e-mails: yao.s.wang@gmail.com, timmy.zhaoqian@gmail.com, dymeng@mail.xjtu.edu.cn}}

%
%

\markboth{Journal of \LaTeX\ Class Files,~Vol.~14, No.~8, August~2015}%
{Shell \MakeLowercase{\textit{et al.}}: Bare Demo of IEEEtran.cls for IEEE Journals}
%



\maketitle

\begin{abstract}
Because of the limitations of matrix factorization, such as losing spatial structure information, the concept of low-rank tensor factorization (LRTF) has been applied for the recovery of a low dimensional subspace from high dimensional visual data.
The low-rank tensor recovery is generally achieved by minimizing the loss function between the observed data and the factorization representation. The loss function is designed in various forms under different noise distribution assumptions, like $L_1$ norm for Laplacian distribution and $L_2$ norm for Gaussian distribution. However, they often fail to tackle the real data which are corrupted by the noise with unknown distribution.
In this paper, we propose a generalized weighted low-rank tensor factorization method (GWLRTF) integrated with the idea of noise modelling. This procedure treats the target data as high-order tensor directly and models the noise by a Mixture of Gaussians, which is called MoG GWLRTF. The parameters in the model are estimated under the EM framework and through a new developed algorithm of weighted low-rank tensor factorization. We provide two versions of the algorithm with different tensor factorization operations, i.e., CP factorization and Tucker factorization.
Extensive experiments indicate the respective advantages of this two versions in different applications and also demonstrate the effectiveness of MoG GWLRTF compared with other competing methods.
\end{abstract}


%
\IEEEpeerreviewmaketitle

\section{Introduction}
\begin{figure}[!t]
\centering
   \includegraphics[width=1.0\linewidth]{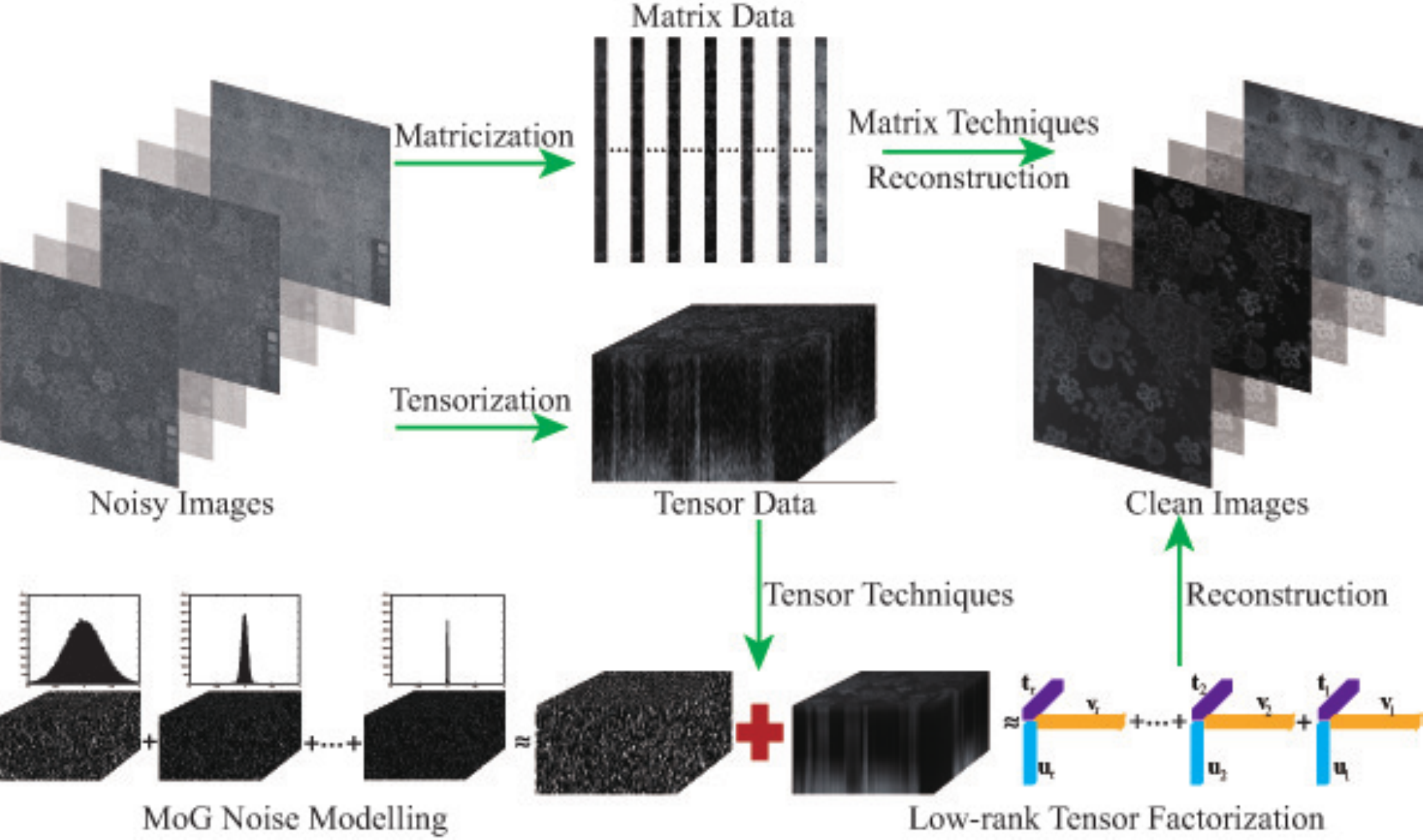}
   \caption{Framework of high-order data processing by matrix techniques and robust tensor factorization method (take CP factorization for illustration). In the initial high-order data representation stage, tensorization better preserves the essential data structure compared with matricization. Furthermore, the integrated MoG model make the tensor factorization more robust to unknown complex noise.} 
\label{fighead}
\end{figure}

%
%
%
%
\IEEEPARstart{T}{he} problem of recovering a low dimensional linear subspace from high dimensional visual data naturally arises in the fields of computer vision, machine learning and statistics, and has drawn increasing attention in the recent years. Typical examples include representation and recognition of faces \cite{wright2009robust,sirovich1987low,turk1991face,belhumeur1997eigenfaces}, structure from motion \cite{tomasi1992shape}, recognition of 3D objects under varying pose \cite{murase1993learning}, motion segmentation \cite{vidal2008multiframe}.
~In such contexts, the data to be analyzed usually can be formulated as high-order tensors, which are natural generalization of vectors and matrices.
Existing approaches, including LRMF and RPCA, proceed by matricizing tensors into matrices and then applying common matrix techniques to deal with tensor problems.
However, as shown in \cite{liu2013}, such matricization procedure fails to exploit the essential tensor structure and often leads to suboptimal procedure.
Figure~\ref{fighead} illustrates the difference between the matrix based method and tensor based method in dealing with the high-order tensor data. The upper row is the matrix based factorization method, which needs to preliminarily matricize the tensor at the cost of losing data structure information; the lower row is our tensor based method which directly factorizes the tensor without destroying the spatial structures.
Given a high-order tensor data, an efficient way to extract the underlying useful information is low-rank tensor factorization (LRTF), which aims to extract low-rank subspaces underlying those vector spaces such that the original tensor can be suitably expressed through reasonably affiliating these subspaces. 

%
Traditionally, there are two definitions of tensor factorizations, i.e., CP factorization and Tucker factorization.

The CP factorization can be viewed as a higher-order generalization of the matrix singular value decomposition \cite{carroll1970analysis,harshman1970foundations,kiers2000towards,Chang1970} and has been widely applied to many fields, including image inpainting~\cite{zhaobayesian,rai2014scalable}, collaborative filtering~\cite{xiong2010temporal}, and data mining~\cite{savas2003analyses}, etc. The idea of CP is to express a tensor as the sum of a finite number of rank-1 tensors. Mathematically, an $N$-order tensor $\mathcal{X} \in \mathbb{R}^{I_1\times I_2\times \cdot\cdot\cdot \times I_N}$, with the integer $I_n$ $(1\le n\le N)$ indicating the dimension of $\mathcal{X}$ along the $n$-th order, is represented in the CP factorization form as
\begin{equation}\label{CP}
   \mathcal{X}=\sum\limits_{d=1}^{r}\bold{u}_{d}\circ \bold{v}_{d}\circ \cdot\cdot\cdot \circ \bold{t}_{d},
\end{equation}
here, $\circ$ denotes the vector outer product and $r$ is assumed to be the rank of the tensor $\mathcal{X}$, denoted as rank($\mathcal{X}$), which is defined as the smallest number of rank-1 tensors~\cite{kruskal1977three,hitchcock1927expression}.

Each element of the tensor has the following form:
\begin{equation}\label{eq_cpele}
   x_{i_1i_2\cdot\cdot\cdot i_N}=\sum\limits_{d=1}^{r}{u_{i_1d}v_{i_2d}\cdot\cdot\cdot t_{i_Nd}},
\end{equation}
where $i_1=1,...,I_1,i_2=1,...,I_2,...,i_N=1,...,I_N.$ The mode (or factor) matrices refer to the combination of the vectors from the rank-1 components, i.e.,
$U=[\bold{u}_{1} \bold{u}_{2} \cdot\cdot\cdot \bold{u}_{r}]$ and likewise for others.

The Tucker factorization is a form of higher-order principal component analysis~\cite{carroll1970analysis,harshman1970foundations,kiers2000towards,tucker1966some} and has been widely used in many applications~\cite{de2004dimensionality,muti2005multidimensional,savas2007handwritten,vasilescu2002multilinear,vasilescu2003multilinear,vlasic2005face,wang2003facial}. It decomposes a tensor into a core tensor multiplied (or transformed) by a matrix along each mode, written as
\begin{equation}\label{eq_Tucker}
\begin{split}
  &\mathcal{X}=\mathcal{G}{{\times }_{1}}U{{\times }_{2}}V \cdot \cdot \cdot {{\times }_{N}}T  \\
    & =\sum\limits_{d_1=1}^{r_1}{\sum\limits_{d_2=1}^{r_2}{\cdot \cdot \cdot \sum\limits_{d_N=1}^{r_N}{{{g}_{d_1d_2 \cdot\cdot\cdot d_N}}{\bold{u}_{d_1}}\circ {\bold{v}_{d_2}}\circ \cdot \cdot \cdot \circ {\bold{t}_{d_N}}}}},
\end{split}
\end{equation}
where $\times$ denotes the $n$-mode matrix product and $\mathcal{G}\in \mathbb{R}^{r_1\times r_2\times \cdot\cdot\cdot \times r_N}$ is the core tensor controlling the interaction between the $N$ mode matrices $U\in \mathbb{R}^{I_1\times r_1},V\in \mathbb{R}^{I_2\times r_2},..,T\in \mathbb{R}^{I_N\times r_N}$.

Elementwise, the Tucker decomposition in Eq.~(\ref{eq_Tucker}) is
\begin{equation}\label{eq_tuckerele}
\begin{split}
  & x_{i_1i_2\cdot\cdot\cdot i_N}\\
    & =\sum\limits_{d_1=1}^{r_1}{\sum\limits_{d_2=1}^{r_2}{\cdot \cdot \cdot \sum\limits_{d_N=1}^{r_N}{{{g}_{d_1d_2 \cdot\cdot\cdot d_N}}{{u}_{i_1d_1}} {{v}_{i_2d_2}} \cdot\cdot\cdot {{t}_{i_Nd_N}}}}}.
\end{split}
\end{equation}
Here the defined $n$-rank of $\mathcal{X}$, denoted as rank$_n(\mathcal{X})$, considers the mode-$n$ rank $r_n$ of tensors. Accordingly, we call $\mathcal{X}$ a rank-$(r_1,r_2,...,r_N)$ tensor.

As mentioned in~\cite{kolda2009tensor}, there exist a number of other tensor factorizations but they are all related to the CP and the Tucker factorization. On account of this, our work for robust low rank tensor factorization is designed based on this two typical factorizations.

It is known that the canonical fit function for the \mbox{LRTF} is based on the Frobenius norm function which assumes the noise to follow a Gaussian distribution. However, for many real data, such as the fMRI neuroimaging data \cite{Friston1996} and the video surveillance data \cite{li2004statistical}, a relative large perturbation in magnitude only affects a relatively small fraction of data points, which often violates the Gaussian assumption and follows a Laplacian distribution instead.

Therefore, it is necessary to consider other loss function that is robust to Laplacian noise. To alleviate this problem, one commonly used strategy is to replace the Frobenius norm function (say, $L_F$ norm) by the $L_1$-type norm \cite{Huang2008, Chi2010}, which is known to be robust to gross Laplacian perturbations. Unfortunately, in many real applications, the noise often exhibits very complex statistical distributions rather than a single purely Gaussian or Laplacian noise~\cite{zhao2015novel}. This motivates us to consider more flexible modeling strategies to tackle such complex noise cases.

Under the framework of low-rank matrix factorization \mbox{(LRMF)}, Meng and De la Torre \cite{meng2013robust} firstly proposed to model the noise as Mixture of Gaussians (MoG). They showed that the MoG model is a universal approximator to any continuous distribution, and hence could be capable of modeling a wider range of noise distributions. Along this line, Zhao et al. \cite{zhao2014robust} further extended the MoG model to deal with robust PCA (RPCA) problem. Extensive experiments on synthetic data, face modeling and background subtraction demonstrated the merits of MoG model.

As such, to share the same light of matrix MoG model, 
we aim to introduce a novel MoG model to the tensor case for the LRTF task to overcome the drawbacks of existing models, which are only optimal for simple Gaussian or Laplacian noise.

The contributions of this paper can be summarized as follows:
(1) As an extension of our last work \cite{chen2016robust}, we propose a generalized low-rank subspace learning approach called generalized weighted low-rank tensor factorization \mbox{(GWLRTF)}, i.e., the GWLRTF-CP and the GWLRTF-Tucker, which both can preserve the essential tensor structure;
(2) For modelling complex noise, MoG is applied to the proposed GWLRTF model called generalized weighted low-rank tensor factorization integrated with MoG \mbox{(MoG GWLRTF)}, i.e., the MoG GWLRTF-CP and the MoG GWLRTF-Tucker;
(3) For solving the proposed model, we propose efficient algorithms to estimate the parameters under the EM framework and through the proposed algorithm of GWLRTF. Our strategy is different from not only the traditional EM algorithm for solving matrix/tensor decomposition models, but also conventional alternative least squares (ALS) techniques for solving other tensor factorization problems;
(4) To further compare the performance between CP factorization and Tucker factorization in different real application problems, a series of synthetic and real data experiments are conducted.
The source codes of our algorithm are published online:
 \emph{http://vision.sia.cn/our\%20team/Hanzhi-homepage/vision-ZhiHan(English).html}.

The rest of this paper is organized as follows. Section 2 describes the notation and common operations used in this paper. In Section 3, a generalized model of tensor factorization integrated with MoG is proposed for reconstructing low-rank tensor from high-order tensor with unknown noise.
In Section 4, the established problem model is solved under the EM framework with new proposed tensor factorization algorithms.
Extensive experiments are conducted on both synthetic data and real image data in Section 5.

\section{Notation and preliminaries}

The notations and common operations used throughout the paper are defined as follows. Scalars are denoted by lowercase letters $(a,b,...)$ and vectors are denoted by bold lowercase letters $\bold{(a,b,...)}$ with elements $(a_i,b_j,...)$. Matrices are represented by uppercase letters $(A,B,...)$ with column vectors $(\bold{a}_{:j},\bold{b}_{:j},...)$ and elements $(a_{ij},b_{ij},...)$. The calligraphic letters $\mathcal{(A,B,...)}$ stand for the the high-order tensors.
We denote an $N$-order tensor as $\mathcal{X}\in \mathbb{R}^{I_1\times I_2\times \cdot\cdot\cdot \times I_N}$, where $I_n (n=1,2,...N)$ is a positive integer. Each element in it is represented as $x_{i_1...i_n...i_N}$, where $1\le i_n\le I_N$.

If tensor $\mathcal{X}$ is a rank-1 tensor, then it can be written as the outer product of $N$ vectors, i.e.,
\begin{equation}\label{rankone}
  \mathcal{X}=\bold{u}\circ \bold{v}\circ \cdot\cdot\cdot \circ \bold{t}.
\end{equation}
Each element of the tensor is the product of the corresponding vector elements which can be represented as:
\begin{equation}\label{rankoneelement}
  x_{i_1i_2\cdot\cdot\cdot i_N}=u_{i_1}v_{i_2}\cdot\cdot\cdot t_{i_N}.
\end{equation}

The slice of an $N$-order tensor is a matrix defined by fixing every index but two. For instance, the slice of a 3-order tensor $\mathcal{X}\in \mathbb{R}^{I\times J\times K}$ has the form: frontal slices $X_{::k}$, lateral slices $X_{:j:}$, horizontal slices $X_{i::}$.
Meanwhile, each order of a tensor is associated with a `mode' and the unfolding matrix of a tensor in each mode is obtained by unfolding the tensor along its corresponding mode. For example, the mode-$n$ unfolding matrix $X_{(n)} \in {{\mathbb{R}}^{I_n\times \prod\limits_{i\ne n}{{{I}_{i}}}}}$ of $\mathcal{X}$, denoted as $X_{(n)}=$unfold$_n(\mathcal{X})$. The inverse operation of the mode-$n$ unfolding is the mode-$n$ folding, represented as $\mathcal{X}=$fold$_n(X_{(n)})$. The mode-$n$ rank $r_n$ of $\mathcal{X}$ is defined as the rank of the mode-$n$ unfolding matrix $X_{(n)}: r_n=$rank$(X_{(n)})$.

The operation of mode-$n$ product of a tensor and a matrix forms a new tensor.
Given tensor $\mathcal{X}\in \mathbb{R}^{I_1\times I_2\times \cdot\cdot\cdot \times I_N}$ and matrix $U \in \mathbb{R}^{J_n \times I_n}$, their mode-$n$ product is calculated by
$\mathcal{X} \times_n U \in \mathbb{R}^{I_1\times\cdot\cdot\cdot \times I_{n-1} \times J_n \times I_{n+1} \times \cdot\cdot\cdot \times I_{N}}$
with element
\begin{equation}\label{modenproductele}
  (\mathcal{X} \times_n U)_{i_1...i_{n-1}j_ni_{n+1}...i_N}=\sum\limits_{i_n}{{x_{i_1...i_n...i_N}}{u_{{j_n{{i_n}}}}}}.
\end{equation}

Given two same-sized tensors $\mathcal{X}, \mathcal{Y} \in \mathbb{R}^{I_1\times I_2\times \cdot\cdot\cdot \times I_N}$, their inner product is defined as:
\begin{equation}\label{tensorproduct}
  \left\langle \mathcal{X},\mathcal{Y} \right\rangle =\sum\limits_{{{i}_{1}}}{\sum\limits_{{{i}_{2}}}{\cdot \cdot \cdot }}\sum\limits_{{{i}_{N}}}{{{x}_{{{i}_{1}}...{{i}_{N}}}}{{y}_{{{i}_{1}}...{{i}_{N}}}}}.
\end{equation}

The Frobenius norm is ${{\left\| \mathcal{X} \right\|}_{F}}=\sqrt{\left\langle \mathcal{X},\mathcal{X} \right\rangle }$. The $l_0$ norm
${{\left\| \mathcal{X} \right\|}_{0}}$ is to calculate the number of non-zero entries in $\mathcal{X}$, and the $l_1$ norm ${{\left\| \mathcal{X} \right\|}_{1}}=\sum\limits_{i_1,...,i_N}\left| {x_{i_1,...,i_N}} \right|$. To be noted, ${{\left\| \mathcal{X} \right\|}_{F}}={{\left\| X_{(n)} \right\|}_{F}}$,${{\left\| \mathcal{X} \right\|}_{0}}={{\left\| X_{(n)} \right\|}_{0}}$, and ${{\left\| \mathcal{X} \right\|}_{1}}={{\left\| X_{(n)} \right\|}_{1}}$ for any $1\le n \le N$.

\section{MoG GWLRTF Model}

In this section, a generalized weighted low-rank tensor factorization model integrated with MoG (MoG GWLRTF) for modelling complex noise is proposed. By applying MoG to model the noise elements of the input tensor, we obtain the log-likelihood optimization objective.

Firstly, taking the noise part (denoted as $\mathcal{E }$) into consideration, the input tensor $\mathcal{X}$
can be represented as:
\begin{equation}\label{eq_dec}
   \mathcal{X}=\mathcal{L}+\mathcal{E },
\end{equation}
where $\mathcal{L}$ denotes the low-rank tensor and the corresponding elementwise form is
\begin{equation}\label{eq_decele}
   x_{i_1i_2\cdot\cdot\cdot i_N}=l_{i_1i_2\cdot\cdot\cdot i_N}+{{\varepsilon }_{i_1i_2\cdot\cdot\cdot i_N}}.
\end{equation}

As MoG has the ability to universally approximate any hybrids of continuous distributions, it is adopted for modeling the unknown noise in the original data. Hence every ${{\varepsilon }_{i_1i_2\cdot\cdot\cdot i_N}}$ follows an MoG and the distribution $p(\varepsilon )$ is defined as:
\begin{equation}\label{eq_gaussiandistribution}
  p(\varepsilon )\sim \sum\limits_{k=1}^{K}{{{\pi }_{k}}\mathcal{N}(\varepsilon |\mu_k,\sigma _{k}^{2})},
\end{equation}
where ${\pi }_{k}$ is the mixing proportion with ${\pi }_{k}\geq0$ and $\sum\limits_{k=1}^{K}
{\pi}_{k}=1$. $\mathcal{N}(\varepsilon |\mu_k,\sigma _{k}^{2})$ denotes the Gaussian distribution with mean $\mu_k$ and variance $\sigma _{k}^{2}$.

Therefore, the element ${x}_{i_1i_2\cdot\cdot\cdot i_N}$ in Eq.~(\ref{eq_decele}) 
follows a MoG distribution with mean $\Lambda_k={l}_{i_1i_2\cdot\cdot\cdot i_N}+\mu_k$
and variance $\sigma _{k}^{2}$. The probability of each element ${x}_{i_1i_2\cdot\cdot\cdot i_N}$ in the input tensor $\mathcal{X}$ can be represented as:
\begin{equation}\label{eq_elementdistribution}
  p({{x}_{i_1i_2\cdot\cdot\cdot i_N}}\left|\right. \Pi, \Lambda, \Sigma )=\sum\limits_{k=1}^{K}{{{\pi }_{k}}\mathcal{N}({x_{i_1i_2\cdot\cdot\cdot i_N}}\left|\Lambda_k, \right.\sigma _{k}^{2})},
\end{equation}
where $\Pi =\left\{ {{\pi }_{1}},{{\pi }_{2}},...,{{\pi }_{K}} \right\},\Lambda=\{{\Lambda_1},{\Lambda_2},...,{\Lambda_K}\},\Sigma =\left\{ {{\sigma }_{1}},{{\sigma }_{2}},...,{{\sigma }_{K}} \right\}$.
Correspondingly, the likelihood of $\mathcal{X}$ can be written as
\begin{equation}\label{eq_likelihood}
 p(\mathcal{X}\left| \Pi, \Lambda, \Sigma  \right.)
  =\prod\limits_{i_1i_2\cdot\cdot\cdot i_N\in \Omega }{\sum\limits_{k=1}^{K}{{{\pi }_{k}}\mathcal{N}({{x}_{i_1i_2\cdot\cdot\cdot i_N}}\left| \Lambda_k, \right.\sigma _{k}^{2})}},
\end{equation}
where $\Omega$ is the index set of the non-missing entries of $\mathcal{X}$.

Our goal is to maximize the logarithmic form of Eq.~(\ref{eq_likelihood}) with respect to the parameters $\Pi, \Lambda, \Sigma$, i.e.
\begin{equation}\label{eq_loglikelihoodold}
\begin{split}
&({\Pi}^*, {\Lambda}^*, {\Sigma}^*)=
\mathop{\argmax}_{\Pi, \Lambda, \Sigma}~{\log}~p(\mathcal{X}\left| \Pi, \Lambda, \Sigma  \right.)\\
 &=\mathop{\argmax}_{\Pi, \Lambda, \Sigma}\sum\limits_{i_1i_2\cdot\cdot\cdot i_N\in \Omega }{\log }\sum\limits_{k=1}^{K}{{{\pi }_{k}}\mathcal{N}({{x}_{i_1i_2\cdot\cdot\cdot i_N}}\left| \Lambda_k, \right.\sigma _{k}^{2})}.\\
\end{split}
\end{equation}

Note that the original problem can be viewed as a Gaussian Scale Mixtures (GSM) with ${\mu_k}$ assumed to be 0, which has been widely used in previous works~\cite{scheunders2007wavelet,wainwright1999scale}.
Therefore, Eq.~(\ref{eq_loglikelihoodold}) can be rewritten as:
\begin{equation}\label{eq_loglikelihood}
\begin{split}
&({\Pi}^*, {\mathcal{L}}^*, {\Sigma}^*)=\mathop{\argmax}_{\Pi, \Lambda, \Sigma}~{\log}~p(\mathcal{X}\left| \Pi, \mathcal{L}, \Sigma  \right.)\\
&=\mathop{\argmax}_{\Pi, \Lambda, \Sigma}~\sum\limits_{i_1i_2\cdot\cdot\cdot i_N\in \Omega }{\log }\sum\limits_{k=1}^{K}{{{\pi }_{k}}\mathcal{N}({{x}_{i_1i_2\cdot\cdot\cdot i_N}}\left| {{l}_{i_1i_2\cdot\cdot\cdot i_N}},\right.\sigma _{k}^{2})}.
\end{split}
\end{equation}

\section{Algorithms under EM framework}

In this section, through assuming a latent variable with higher dimension, we solve the above problem iteratively under the EM framework.
We also design algorithms to solve the generalized weighted low-rank tensor factorization model for updating the low-rank tensor.

EM algorithm \cite{dempster1977maximum} is proven to be effective for solving the maximization problem of the log-likelihood function.
Therefore, for solving Eq.~(\ref{eq_loglikelihoodold}), we assume a higher dimensional latent variable under the EM framework.

In the model, 
the factorized low-rank tensor components are shared by all the clusters of MoG and the mean for each cluster of the standard EM algorithm is represented by them. Thus our proposed algorithm will iterate between computing responsibilities of all Gaussian components ({\bf E Step}) and maximizing the parameters $\Pi, \Sigma$ and the low-rank tensor $\mathcal{L}$ in the model ({\bf M Step}).

{\bf\emph{ E Step}:} A latent variable $z_{i_1i_2\cdot\cdot\cdot i_N k}$ is assumed in the model, with ${z}_{i_1i_2\cdot\cdot\cdot i_N k}\in \{0,1\}$ and $\sum\limits_{k=1}^{K}
{z}_{i_1i_2\cdot\cdot\cdot i_N k}=1$, representing the assigned value of the noise ${\varepsilon }_{i_1i_2\cdot\cdot\cdot i_N}$ to each component of the mixture. Here we denote $Z=\{z_{i_1i_2\cdot\cdot\cdot i_N k}|i_1=1,2,...,I_1; i_2=1,2,...,I_2;...; i_N=1,2,...,I_N; k=1,2,...,K\}$.
The posterior responsibility of the $k$-th mixture for generating the noise of ${x_{i_1i_2\cdot\cdot\cdot i_N}}$ can be calculated by
\begin{equation}\label{eq_latent}
\begin{split}
&E({{z}_{i_1i_2\cdot\cdot\cdot i_N k}})={{\gamma }_{i_1i_2\cdot\cdot\cdot i_N k}}\\
&=\frac{{{\pi }_{k}}\mathcal{N}({{x}_{i_1i_2\cdot\cdot\cdot i_N}}\left| {{l}_{i_1i_2\cdot\cdot\cdot i_N}}, \right.\sigma _{k}^{2})}{\sum\limits_{k=1}^{K}{{{\pi }_{k}}\mathcal{N}({{x}_{i_1i_2\cdot\cdot\cdot i_N}}\left| {{l}_{i_1i_2\cdot\cdot\cdot i_N}}, \right.\sigma _{k}^{2})}}.
\end{split}
\end{equation}

The {\bf M step} maximizes the upper bound given by the {\bf E step} with regard to $\mathcal{L},\Pi,\Sigma$:
\begin{equation}\label{eq_Mstep}
  \begin{split}
  &{{E}_{Z}} p(\mathcal{X},Z|\mathcal{L},\Pi ,\Sigma)=\sum\limits_{i_1i_2\cdot\cdot\cdot i_N\in \Omega }\sum\limits_{k=1}^{K}{{\gamma }_{i_1i_2\cdot\cdot\cdot i_N k}}{(}log{{\pi }_{k}}\\
  &-log\sqrt{2\pi {{\sigma }_{k}}}-\frac {( {{x}_{i_1i_2\cdot\cdot\cdot i_N}}-{{l}_{i_1i_2\cdot\cdot\cdot i_N}}{{)}^{2}}}{2\pi \sigma _{k}^{2}} {)}.\\
  \end{split}
\end{equation}

This maximization problem can be solved by alternatively updating the MoG parameters $\Pi,\Sigma$ and the factorized 
components of low-rank tensor $\mathcal{L}$ as follows:

{\bf\emph{ M Step}}\emph{ to update} \emph{$\Pi, \Sigma$}: The closed-form updates for the MoG parameters are:
\begin{equation}\label{eq_PiSigma}
  \begin{split}
  &{{m}_{k}}=\sum\limits_{i_1,i_2,\cdot\cdot\cdot, i_N}{{{\gamma }_{i_1i_2\cdot\cdot\cdot i_Nk}}},{{\pi }_{k}}=\frac{{{m}_{k}}}{\sum\limits_{k}{m}_{k}}, \\
  &\sigma _{k}^{2}=\frac{1}{{{m}_{k}}}\sum\limits_{i_1i_2\cdot\cdot\cdot i_N}{{{\gamma }_{i_1i_2\cdot\cdot\cdot i_Nk}}({{x}_{i_1i_2\cdot\cdot\cdot i_N}}-{{l}_{i_1i_2\cdot\cdot\cdot i_N}}{{)}^{2}}}.
  \end{split}
\end{equation}

{\bf \emph{M Step}}\emph{ to update $\mathcal{L}$}: Re-write Eq.~(\ref{eq_Mstep}) only with regard to the unknown $\mathcal{L}$
as follows:
\begin{equation}\label{eq_UVT}
 \begin{split}
 &\sum\limits_{i_1,i_2,\cdot\cdot\cdot, i_N\in \Omega }{\sum\limits_{k=1}^{K}{{{\gamma }_{i_1i_2\cdot\cdot\cdot i_Nk}}(-\frac{({{x}_{i_1i_2\cdot\cdot\cdot i_N}}-{{l}_{i_1i_2\cdot\cdot\cdot i_N}}{{)}^{2}}}{2\pi \sigma _{k}^{2}})}}\\
 &=-\sum\limits_{i_1,i_2,\cdot\cdot\cdot, i_N\in \Omega }{\sum\limits_{k=1}^{K}
 ({\frac{{{\gamma }_{i_1i_2\cdot\cdot\cdot i_Nk}}}{2\pi \sigma _{k}^{2}})({{x}_{i_1i_2\cdot\cdot\cdot i_N}}-{{l}_{i_1i_2\cdot\cdot\cdot i_N}}{{)}^{2}}}}\\
 &=-{{\left\| \mathcal{W}\odot (\mathcal{X}-\mathcal{L}) \right\|}^{2}_{{{L}_{F}}}}.
 \end{split}
\end{equation}
Here $\odot$ denotes the Hadamard product (component-wise multiplication) and the element ${w_{i_1i_2\cdot\cdot\cdot i_N}}$ of $\mathcal{W} \in {\mathbb{R}^{I_1\times I_2\times \cdot\cdot\cdot \times I_N}}$ is
\begin{equation}\label{eq_innerW}
  {{w}_{i_1i_2\cdot\cdot\cdot i_N}}=\left\{
  \begin{split}
  & \begin{matrix}
   \sqrt{\sum\limits_{k=1}^{K}{\frac{{{\gamma }_{i_1i_2\cdot\cdot\cdot i_Nk}}}{2\pi \sigma _{k}^{2}}}}, & i_1,i_2,\cdot\cdot\cdot, i_N\in \Omega   \\
\end{matrix} \\
 & \begin{matrix}
   0, & i_1,i_2,\cdot\cdot\cdot, i_N\notin \Omega.   \\
\end{matrix} \\
\end{split} \right.
\end{equation}

The whole MoG GWLRTF optimization process is summarized in Algorithm \ref{MoG-LRTF}.

\begin{algorithm}[!htb]
\caption{(Algorithm for MoG GWLRTF)}
\label{MoG-LRTF}
\begin{algorithmic}[1]
\REQUIRE ~~
 the original data represented in tensor form $\mathcal{X}\in \mathbb{R}^{I_1\times I_2\times \cdot\cdot\cdot \times I_N}$\\
\ENSURE
the recovered low-rank tensor $\mathcal{L}$
\STATE
Initialize ${\Pi},{\Sigma}, \mathcal{L}$, MoG number $K$, small threshold $\epsilon$.\\
\WHILE{not converged}
\STATE {\bf E Step}: \\
Evaluate ${{\gamma}_{i_1i_2\cdot\cdot\cdot i_Nk}}$ by Eq.~(\ref{eq_latent}).
\STATE {\bf M Step} for ${\Pi},{\Sigma}$: \\
Evaluate $\pi_k,\sigma_{k}^{2}$ by Eq.~(\ref{eq_PiSigma})
\STATE {\bf M Step} for $\mathcal{L}$: \\
Evaluate $\mathcal{L}$ 
 by solving the GWLRTF model \\
 ${\min\limits_{\mathcal{L}}}{{\left\| \mathcal{W}\odot (\mathcal{X}-\mathcal{L}) \right\|}^{2}_{{{L}_{F}}}}$, \\
  where $\mathcal{W}$ is calculated by Eq.~(\ref{eq_innerW}).\\
\ENDWHILE
\end{algorithmic}
\end{algorithm}

In the {\bf M Step}, 
 $\mathcal{L}$ is evaluated by solving the GWLRTF model ${\min\limits_{\mathcal{L}}}{{\left\| \mathcal{W}\odot (\mathcal{X}-\mathcal{L}) \right\|}^{2}_{{{L}_{F}}}}$.
Here, we introduce the two typical factorizations to the GWLRTF and the corresponding algorithms are given in detail in the following parts.

\subsection{Weighted low-rank tensor CP factorization (GWLRTF-CP)}

The GWLRTF model of a 3-order tensor $\mathcal{X} \in \mathbb{R}^{I\times J\times K}$ in the form of CP factorization can be written as
 \begin{equation}\label{eq_LRTF}
  {\min_{U, V, T}}{{\left\| \mathcal{W}\odot (\mathcal{X}-\sum\limits_{d=1}^{r}{{\bold{u}_{:d}}\circ {\bold{v}_{:d}}\circ {\bold{t}_{:d}}}) \right\|}_{{{L}_{F}}}},
 \end{equation}
 where
 \begin{equation*}
   \mathcal{L}=\sum\limits_{d=1}^{r}{{\bold{u}_{:d}}\circ {\bold{v}_{:d}}\circ {\bold{t}_{:d}}}
 \end{equation*}
 and $U \in \mathbb{R}^{I\times r}, V \in \mathbb{R}^{J\times r}, T \in \mathbb{R}^{K\times r}$ are mode matrices with rank $r$. $\mathcal{W}\in \mathbb{R}^{I\times J\times K}$ is the weighted tensor which is composed by the standard variance of the input tensor elements.

 Because of the effectiveness and implementation convenience of ALS, we adopt its idea to update $U, V, T$ of the tensor one at a time.

 Suppose $\bold{I_{1}^{0},...,I_{n}^{0}}\in {{\mathbb{R}}^{w\times h}}$ are data matrices. In order to stack each of the above matrix as a vector, we define the operator $vec:{{\mathbb{R}}^{\text{w}\times h}}\to {{\mathbb{R}}^{wh}}$.

For each slice of the higher-order tensor, it can be viewed as a linear combination of the corresponding slices of all the rank-1 tensors. Different from other methods for solving the problem of LRTF, we stack each frontal slice of the higher-order tensor as a vector of a new matrix denoted as $M_F$. Correspondingly, the vectorized horizontal slices and lateral slices are represented as ${M}_H$ and ${M}_L$, respectively.

Firstly we have
\begin{equation}\label{Xnew}
  \mathcal{X}^{weight}=\mathcal{W}\odot\mathcal{X}.
\end{equation}
Then taking term ${T}$ as an example, the vectorized frontal slice ${M}_F$ of the higher-order tensor can be written as follows:
\begin{equation}\label{surfaceTensorT}
 \begin{split}
  {M}_F=[vec(X^{weight}_{::1})|...|vec(X^{weight}
  _{::K})]
   \in \mathbb{R}^{IJ\times K}.
  \end{split}
 \end{equation}

For the $i$-th frontal slice of the higher-order tensor, the vectorized corresponding slices of all the rank-1 tensors can be viewed as the $i$-th element of the cell $F$ which can be represented as:
\begin{equation}\label{eq_surfaceT}
 \begin{split}
 F_i=[vec(W_{::i}\odot&(\bold{u}_{:1}^{old}\circ \bold{v}_{:1}^{old}))|...\\
 &|vec(W_{::i}\odot(\bold{u}_{:r}^{old}\circ \bold{v}_{:r}^{old}))] \in \mathbb{R}^{IJ\times r}.
 \end{split}
 \end{equation}

Then the $i$-th vector of term $T$ can be updated as follows:
\begin{equation}\label{eq_T}
    {T}_{i:}^{new}=(F_i^{\dagger}{{M}_F}_{:i})^ \mathrm{ T }  \in \mathbb{R}^{1\times r},
 \end{equation}
where $A^{\dagger}$ represents the pseudo-inverse matrix of matrix $A$, and ${B}^ \mathrm{ T } $ denotes the transposed matrix of matrix ${B}$.

Similarly, we have the term ${V}$ and ${U}$ updated as following:
\begin{equation}\label{surfaceTensorV}
 \begin{split}
  {M}_L=[vec(X^{weight}_{:1:})|...|vec(X^{weight}
  _{:J:})]\in \mathbb{R}^{IK\times J},
  \end{split}
 \end{equation}
\begin{equation}\label{eq_surfaceV}
 \begin{split}
 {L}_i=[vec({W}_{:i:}\odot&(\bold{t}_{:1}^{new}\circ \bold{u}_{:1}^{old}))|...\\
 &|vec({W}_{:i:}\odot(\bold{t}_{:r}^{new}\circ \bold{u}_{:r}^{old}))] \in \mathbb{R}^{IK\times r},
 \end{split}
 \end{equation}
 \begin{equation}\label{eq_V}
    {V}_{i:}^{new}=({L}_i^{\dagger}{{M}_L}_{:i})^ \mathrm{ T }   \in \mathbb{R}^{1\times r}.
 \end{equation}
 \begin{equation}\label{surfaceTensorU}
 \begin{split}
 {M}_H=[vec(X^{weight}_{1::})|...|vec(X^{weight}
  _{I::})] \in \mathbb{R}^{JK\times I},
  \end{split}
 \end{equation}
 \begin{equation}\label{eq_surfaceU}
 \begin{split}
 {H}_i=[vec({W}_{i::}\odot&(\bold{v}_{:1}^{new}\circ \bold{t}_{:1}^{new}))|...\\
 &|vec({W}_{i::}\odot(\bold{v}_{:r}^{new}\circ \bold{t}_{:r}^{new}))] \in \mathbb{R}^{JK\times r},
 \end{split}
 \end{equation}
 \begin{equation}\label{eq_U}
    {U}_{i:}^{new}=({H}_i^{\dagger}{{M}_H}_{:i})^ \mathrm{ T }   \in \mathbb{R}^{1\times r}.
 \end{equation}

 The whole optimization process is summarized in Algorithm \ref{LRTF}.
 \begin{algorithm}[htb]
 \caption{(GWLRTF-CP)}
 \label{LRTF}
 \begin{algorithmic}[1]
 \REQUIRE ~ The input tensor $\mathcal{X}\in \mathbb{R}^{I\times J\times K}$, initialized tensor factors ${U, V, T}$, weighted
 tensor $\mathcal{W}$, number of iteration and the threshold $\epsilon$.
 \ENSURE ${U, V, T}$.
\WHILE{not converged}
 \STATE update ${T}$ with Eq.~(\ref{surfaceTensorT}),~(\ref{eq_surfaceT}),~(\ref{eq_T});
 \STATE update ${V}$ with Eq.~(\ref{surfaceTensorV}),~(\ref{eq_surfaceV}),~(\ref{eq_V});
 \STATE update ${U}$ with Eq.~(\ref{surfaceTensorU}),~(\ref{eq_surfaceU}),~(\ref{eq_U}).
\ENDWHILE
 \end{algorithmic}
 \end{algorithm}

\subsection{Weighted low-rank tensor Tucker factorization (GWLRTF-Tucker)}

By applying the Tucker factorization to the low-rank tensor $\mathcal{L}$ in the GWLRTF model, we obtain the following GWLRTF-Tucker model
\begin{equation}\label{eq_WeiTucker}
{\min_{\mathcal{G}, U, V,.., T}}{{\left\| \mathcal{W}\odot (\mathcal{X}-\mathcal{G}{{\times }_{1}}U{{\times }_{2}}V \cdot \cdot \cdot {{\times }_{N}}T) \right\|}_{{{L}_{F}}}}.
\end{equation}

Through coordinate-wisely separating the original optimization problem into solving a sequence of scalar minimization subproblems, coordinate descent has exhibited its effectiveness in dealing with the convex optimization problems~\cite{meng2015robust,friedman2007pathwise,friedman2010regularization}. Based on this, we aim to coordinate-wisely optimize each entry of $U,V, . . . ,T$ and $\mathcal{G}$ in Eq. (\ref{eq_WeiTucker}).

{\bf Update the mode matrices $U,V,...,T$}:
Firstly, we reformulate Eq. (\ref{eq_WeiTucker}) in the form of minimizing the function against only one of the unknown mode matrices (take $H$ for example) at a time with others fixed as
\begin{equation}\label{eq_c1}
     {\left\| \mathcal{W}\odot (\mathcal{X}-\mathcal{D}{{\times }_{n}}H) \right\|}_{L_F},
\end{equation}
where
\begin{equation}\label{eq_c11}
  \mathcal{D}=\mathcal{G}{{\times }_{1}}U{{\times }_{2}}V\cdot \cdot \cdot {{\times }_{n-1}}F{{\times }_{n+1}}K\cdot \cdot \cdot {{\times }_{N}}T.
\end{equation}

Unfolding the tensors along mode-$n$, it can be reformulated as the following sub-problem
\begin{equation}\label{eq_c2}
\begin{split}
   & {{\left\| W_{(n)}\odot ({{X}_{(n)}}-H{{D}_{(n)}}) \right\|}_{{{L}_{F}}}} \\
    & ={{\left\| W_{(n)}\odot ({{X}_{(n)}}-\sum\limits_{j=1}^{{{r}_{n}}}{{{\bold{h}}_{:j}}{{{\bold{d}}_{:j}^{\mathrm{ T }}}}}) \right\|}_{{{L}_{F}}}}\\
& ={{\left\| W_{(n)}\odot (E-{{\bold{h}}_{:k}}{{{\bold{d}}_{:k}^{\mathrm{ T }}}}) \right\|}_{{{L}_{F}}}},
\end{split}
\end{equation}
where
\begin{equation}\label{eq_c31}
  E={{X}_{(n)}}-\sum\limits_{j\ne k}{{{\bold{h}}_{:j}}{{{\bold{d}}_{:j}^{\mathrm{ T }}}}}.
\end{equation}

Then, the original problem Eq. (\ref{eq_WeiTucker}) are separated into the following single-scalar parameter optimization sub-problems
\begin{equation}\label{eq_c3}
\begin{split}
&{\min_{h_{ik}}} {{\left\| {\bold{w}}_{k:}\odot ({\bold{e}}_{k:}- {\bold{d}}_{k:}{{h}_{ik}} ) \right\|}_{{{L}_{F}}}}\\
&={\min_{h_{ik}}} {{\left\| {\bold{w}}_{k:}\odot {\bold{e}}_{k:}-{\bold{w}}_{k:}\odot {\bold{d}}_{k:}{{h}_{ik}} \right\|}_{{{L}_{F}}}}.
  \end{split}
\end{equation}

{\bf Update the core tensor $\mathcal{G}$}:
Likewise, for the equivalent formulation of Eq. (\ref{eq_WeiTucker})
\begin{equation}\label{eq_WeiTuckerZ}
{{\left\| \mathcal{W}\odot (\mathcal{X}-\sum\limits_{d_1=1}^{r_1}{{\cdot \cdot \cdot \sum\limits_{d_N=1}^{r_N}{{{g}_{d_1d_2 \cdot\cdot\cdot d_N}}{\bold{u}_{d_1}}\circ {\bold{v}_{d_2}}\circ \cdot \cdot \cdot \circ {\bold{t}_{d_N}}}}}) \right\|}_{{{L}_{F}}}},
\end{equation}
it can be rewritten as
\begin{equation}\label{eq_WeiTuckerZ2}
  {\left\| \mathcal{W}\odot (\mathcal{E}-g_{k_1...k_N}\mathcal{U})\right\|}_{L_F},
\end{equation}
where
\begin{equation*}\label{eq_WeiTuckerZ21}
  \mathcal{E}=\mathcal{X}-\sum\limits_{d_1\ne k_1,...,d_N\ne k_N}{{g}_{d_1d_2 \cdot\cdot\cdot d_N}}{\bold{u}_{d_1}}\circ {\bold{v}_{d_2}}\circ \cdot \cdot \cdot \circ{\bold{t}_{d_N}},
\end{equation*}
\begin{equation*}\label{eq_WeiTuckerZ21}
  \mathcal{U}={\bold{u}_{k_1}}\circ {\bold{v}_{k_2}}\circ \cdot \cdot \cdot \circ{\bold{t}_{k_N}}.
\end{equation*}

Here we denote $\bold{w}=vec(\mathcal{W})$, $\bold{e}=vec(\mathcal{E})$, $\bold{u}=vec(\mathcal{U})$, then the optimization of Eq. (\ref{eq_WeiTuckerZ}) can be obtained by minimizing the following sub-problems
\begin{equation}\label{eq_WeiTuckerZ3}
\begin{split}
 & \min_{g_{k_1...k_N}}{\left\| \bold{w} \odot (\bold{e}-\bold{u}g_{k_1...k_N})\right\|}_{L_F}\\
  &=\min_{g_{k_1...k_N}} {\left\| \bold{w} \odot \bold{e}- \bold{w} \odot  \bold{u}g_{k_1...k_N})\right\|}_{L_F}.
  \end{split}
\end{equation}

The solutions of Eqs. (\ref{eq_c3}) and (\ref{eq_WeiTuckerZ3}) can be exactly obtained by ALS. The whole optimization process is summarized in Algorithm \ref{LRTF-tucker}.
 \begin{algorithm}[htb]
 \caption{(GWLRTF-Tucker)}
 \label{LRTF-tucker}
 \begin{algorithmic}[1]
 \REQUIRE ~ The input tensor $\mathcal{X} \in \mathbb{R}^{I_1\times I_2\times \cdot\cdot\cdot \times I_N}$, initialized tensor factors $U, V,..., T$ and $\mathcal{G}$, weighted
 tensor $\mathcal{W}$, number of iteration and the threshold $\epsilon$.
 \ENSURE ${U, V,.., T,\mathcal{G}}$.
 \WHILE {not converged}
 \STATE update the entries of mode matrices ${U,V,...,T}$ by solving Eq.~(\ref{eq_c3});
 \STATE update each entry of the core tensor $\mathcal{G}$ with Eq.~(\ref{eq_WeiTuckerZ3});
 \ENDWHILE
 \end{algorithmic}
 \end{algorithm}

\section{Experiments}
\begin{table*}[t]
\footnotesize
 \caption{Predictive performance of competing methods with varied missing rate. In each experiments, the best performance is highlighted in bold and the second is underlined.}
 \label{table1}
 \centering
\begin{tabular}{c|ccccccccc}
\hline
         &    & MoG LRMF & HaLRTC      &LRTA  & PARAFAC& MSI DL &CWM LRTF   &MoG GWLRTF-CP & MoG GWLRTF-Tucker\\
  \hline
         &E1  & 0.08 & 2.55e+02      &4.40e+02  & 4.38e+02& 2.98e+02 &3.51e+02   &\underline{1.61e-08}& \textbf{2.31e-12} \\
  $20\%$ &E2      & 1.08e-04 & 6.50e+04     &6.47e+02  & 6.47e+02& 3.70e+02 &3.93e+02  &\underline{6.82e-19}& \textbf{1.43e-26}   \\
         &E3  & 0.09 & 2.68e+04 &6.07e+02  &5.95e+02& 4.10e+02&4.63e+02  &\underline{2.03e-08}& \textbf{ 2.92e-12 } \\
         &E4  & 0.57 & 3.76e+06 &9.62e+02  & 9.48e+02& 5.82e+02&5.34e+02  &\underline{8.85e-19}&  \textbf{1.85e-26} \\
\hline
         &E1   & 1.24&  2.55e+02   &5.28e+02 & 5.30e+02& 3.58e+02 &4.38e+02  &\underline{8.84e-09} & \textbf{ 1.10e-12}  \\
 $40\%$  &E2   & 0.02 &  6.50e+04   &1.24e+03 & 1.23e+03& 7.95e+02 &8.07e+02 &\underline{3.18e-19} &  \textbf{3.82e-27}    \\
         &E3 & 7.36e+02 & 5.20e+04  &9.93e+02  & 9.78e+02& 6.54e+02 &8.00e+02  &\underline{1.90e-08} & \textbf{ 2.07e-12}  \\
         &E4 & 1.41e+02 & 7.06e+06   &2.28e+03 & 2.22e+03& 1.50e+03 &1.56e+03 &\underline{9.63e-19} &\textbf{ 8.95e-27}  \\
\hline
         &E1   &5.01 &  2.55e+02   &7.11e+02  & 6.63e+02& 4.85e+02 &5.57e+02  &\underline{2.49e-07} & \textbf{ 8.84e-07}\\
$60\%$   &E2    & 0.66     &  6.50e+04  &2.34e+03  & 2.21e+03& 1.67e+03 &1.90e+03&\underline{2.77e-16 } & \textbf{ 3.59e-27 }  \\
         &E3 & 1.51e+04 &  7.89e+04  &1.86e+03 & 1.81e+03& 1.25e+03 &1.73e+03 &\underline{1.12e-06} & \textbf{ 2.48e-12 }  \\
         &E4  & 1.57e+03 & 1.09e+07  &6.26e+03  & 6.21e+03& 4.32e+03 &6.22e+03  &\underline{4.21e-15}& \textbf{ 1.24e-26 } \\
\hline
\end{tabular}
\end{table*}
\begin{table*}[t]
\footnotesize
\caption{Reconstruction performance of competing methods with unknown noise. In each experiments, the best performance is highlighted in bold and the second is underlined.}
\label{table2}
\centering
\begin{tabular}{c|cccccccccc}
\hline
         &  & MoG LRMF & HaLRTC      &LRTA  & PARAFAC& MSI DL &CWM LRTF &MoG GWLRTF-CP&MoG GWLRTF-Tucker\\
  \hline
         &E1    & \underline{32.3} & 2.55e+02   &6.31e+02  & 6.41e+02& 4.42e+02 &4.69e+02  &54.2& \textbf{14.7}   \\
Gaussian &E2 &  \underline{2.44}     & 6.50e+04    &1.55e+03  & 1.57e+03& 1.17e+03 &7.04e+02  &5.84&  \textbf{1.62 }   \\
Noise    &E3  & \textbf{11.4}&1.05e+05  &8.59e+02  & 8.53e+02& 6.02e+02 &6.09e+02 &29.2&  \underline{ 14.8 }   \\
         &E4 &  72.0    & 1.41e+07     &2.10e+03  & 2.07e+03& 1.53e+03 &8.81e+02  &\underline{1.52}&  \textbf{0.647}     \\
\hline
         &E1  & \textbf{4.20e+02}& 5.10e+02   &1.05e+03  & 9.63e+02& 7.70e+02 &8.86e+02  &6.93e+02& \underline{4.36e+02 }   \\
Sparse   &E2  &  \textbf{4.96e+02} & 1.30e+05   &2.96e+03  & 2.63e+03& 2.24e+03 &2.47e+03 &1.42e+03& \underline{1.25e+03}  \\
Noise    &E3&  5.25e+03 & 1.02e+05  &1.02e+03 & 1.05e+03& 7.47e+02 &8.41e+02   &\textbf{5.10e+02}& \underline{6.82e+02} \\
         &E4  & \underline{1.04e+03} & 1.33e+07  &2.17e+03  & 2.34e+03& 1.64e+03&1.77e+03 &\textbf{4.33e+02}& 1.30e+03 \\
\hline
         &E1  & \underline{4.63e+02} & 5.10e+02   &1.19e+03 & 1.15e+03& 8.17e+02 &1.07e+03  &6.68e+02& \textbf{4.46e+02 }\\
Mixture  &E2 & \textbf{6.05e+02} & 1.30e+05  &3.90e+03  & 3.70e+03& 2.58e+03 &3.31e+03 &1.37e+03& \underline{1.29e+03  } \\
Noise    &E3 &  1.23e+04 & 1.06e+05 &1.13e+03  & 1.17e+03& 7.85e+02 &1.10e+03   &\textbf{4.59e+02}& \underline{4.62e+02} \\
         &E4  &1.46e+03 & 1.50e+07  &2.84e+03 & 2.94e+03& 1.93e+03 &2.91e+03 &\textbf{3.83e+02}& \underline{5.72e+02 } \\
\hline
\end{tabular}
\end{table*}

In this section, we conduct extensive experiments on both synthetic data and real applications to validate the effectiveness of the proposed MoG GWLRTF, compared with MoG LRMF~\cite{meng2013robust}, HaLRTC~\cite{liu2013}, LRTA~\cite{renard2008denoising}, PARAFAC~\cite{liu2012denoising}, MSI DL~\cite{peng2014decomposable}, CWM LRTF~\cite{meng2015robust}. Specifically, MoG GWLRTF-Tucker and MoG GWLRTF-CP are also demonstrated to further compare the performance of CP factorization and Tucker factorization in different applications.
The synthetic experiments are designed to quantitatively assess our methods from: i) predictive performance over missing entries given an incomplete tensor data; \mbox{ii)} reconstruction performance given a both incomplete and noisy tensor data. Real data applications, i.e., single RGB image reconstruction, face modeling, multispectral image recovery and real hyperspectral image restoration, are further conducted to validate the effectiveness of the proposed algorithms.

\subsection{Synthetic Experiments}

\begin{figure*}[t]
\begin{center}
   \includegraphics[width=1.0\linewidth]{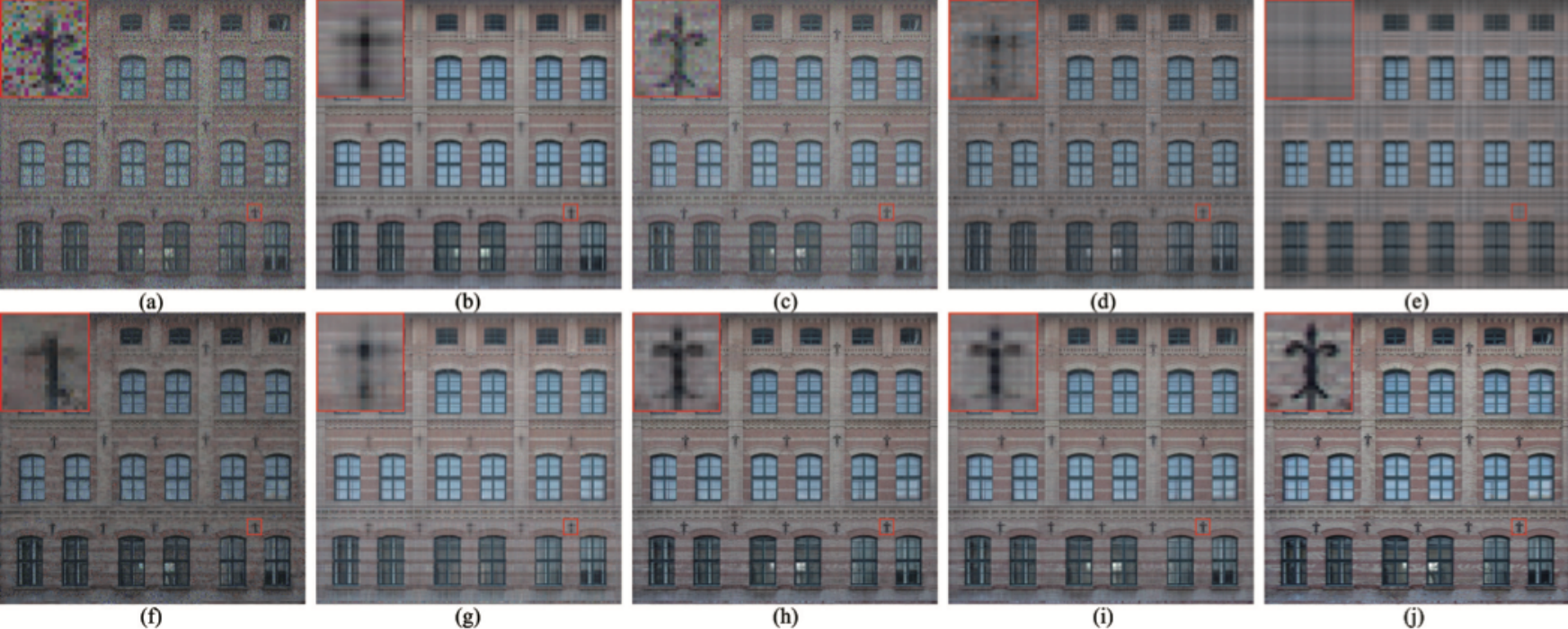}
\end{center}
   \caption{Facade with small mixture noise. (a) Noisy image. (b)MoG LRMF. (c)HaLRTC. (d)LRTA. (e)PARAFAC. (f)MSI DL. (g)CWM LRTF. (h)MoG GWLRTF-CP. (i)MoG GWLRTF-Tucker. (j) Original image.}
\label{fig_facadelessnoise}
\end{figure*}
\begin{figure*}[t]
\begin{center}
   \includegraphics[width=1.0\linewidth]{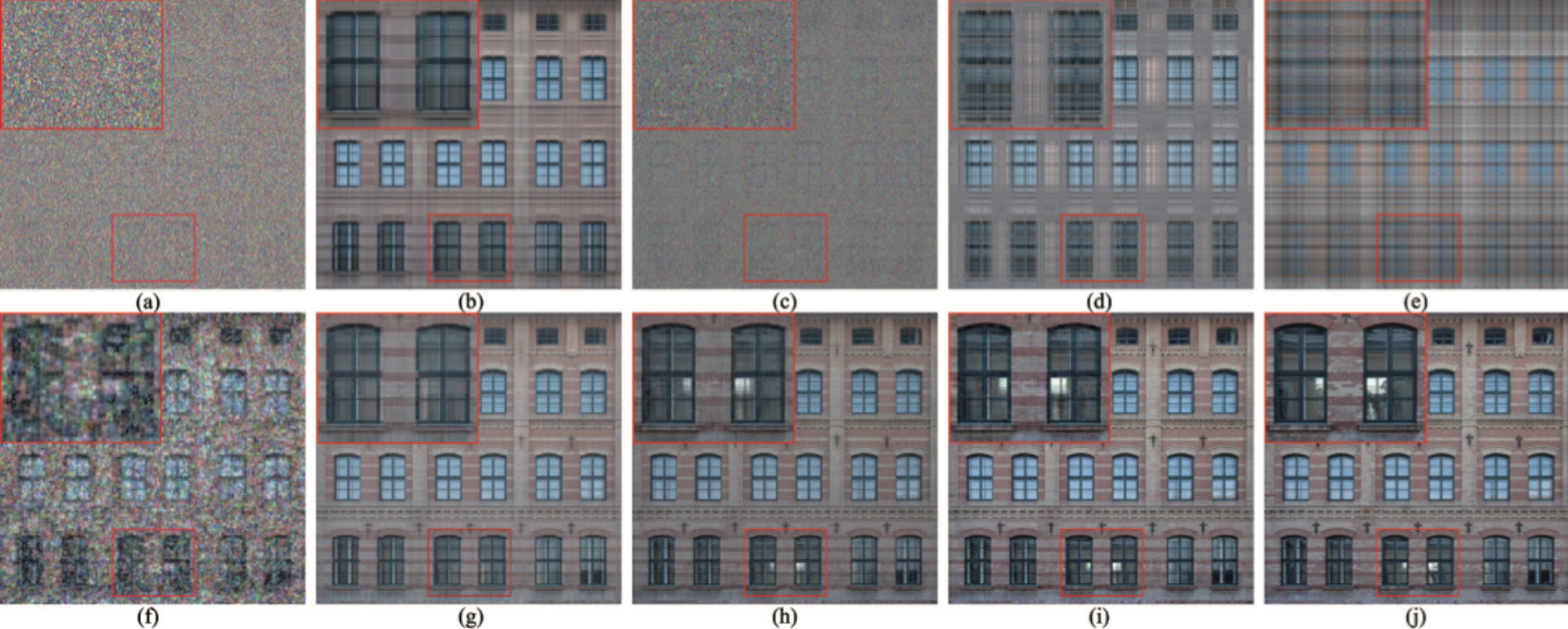}
\end{center}
   \caption{Facade with mixture noise. (a) Noisy image. (b)MoG LRMF. (c)HaLRTC. (d)LRTA. (e)PARAFAC. (f)MSI DL. (g)CWM LRTF. (h)MoG GWLRTF-CP. (i)MoG GWLRTF-Tucker. (j) Original image.}
\label{fig_facade}
\end{figure*}
\begin{table*}[!t]
\caption{Facade reconstruction performance of competing methods with mixture noise. In each experiments, the best performance is highlighted in bold and the second is underlined.}
\label{table3}
\footnotesize
\centering
\begin{tabular}{c|ccccccccc}
\hline
     Facade    &    &  MoG LRMF & HaLRTC      &LRTA  & PARAFAC& MSI DL &CWM LRTF   &MoG GWLRTF-CP &MoG GWLRTF-Tucker\\
  \hline
               &PSNR&   24.34 & 23.43    &13.59   &13.37   &13.53    &24.80   &\underline{25.65}  &   \textbf{ 25.67  }      \\
small        &RSE &    0.1169  & 0.1298  &0.4026  &0.4129  &0.4062   &0.1109  &  \underline{0.1005}&\textbf{0.1003}   \\
noise      &FSIM&    0.8954  & 0.9407  &0.8318  &0.7402  &0.8258   &0.9435&  \textbf{0.9539}& \underline{0.9454}  \\

\hline
          &PSNR &   22.18  &   14.20   &18.51   &16.95   &16.71    &22.82  &\underline{23.69}&   \textbf{ 24.46 }       \\
mixture     &RSE &  0.1499   &   0.3755 &0.2287 &0.2737  &0.2817   &0.1393 & \underline{0.1260}&   \textbf {0.1127}     \\
  noise     &FSIM &   0.8525  &   0.6003 &0.7667 &0.7101  &0.7310   &0.9117 &\underline{0.9268}& \textbf{ 0.9310  }      \\
\hline
\end{tabular}
\end{table*}

\begin{figure*}[!t]
\centering
   \includegraphics[width=1.0\linewidth]{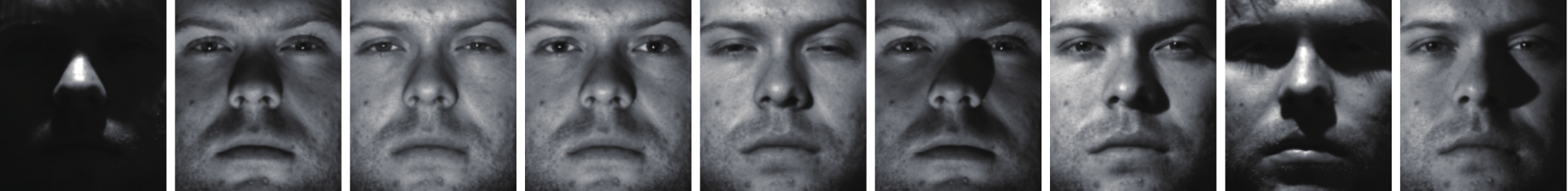}
   \caption{Sampling images of the first people under 9 illuminations.}
\label{fig_oriface}
\end{figure*}
\begin{figure*}[!t]
\centering
   \includegraphics[width=1.0\linewidth]{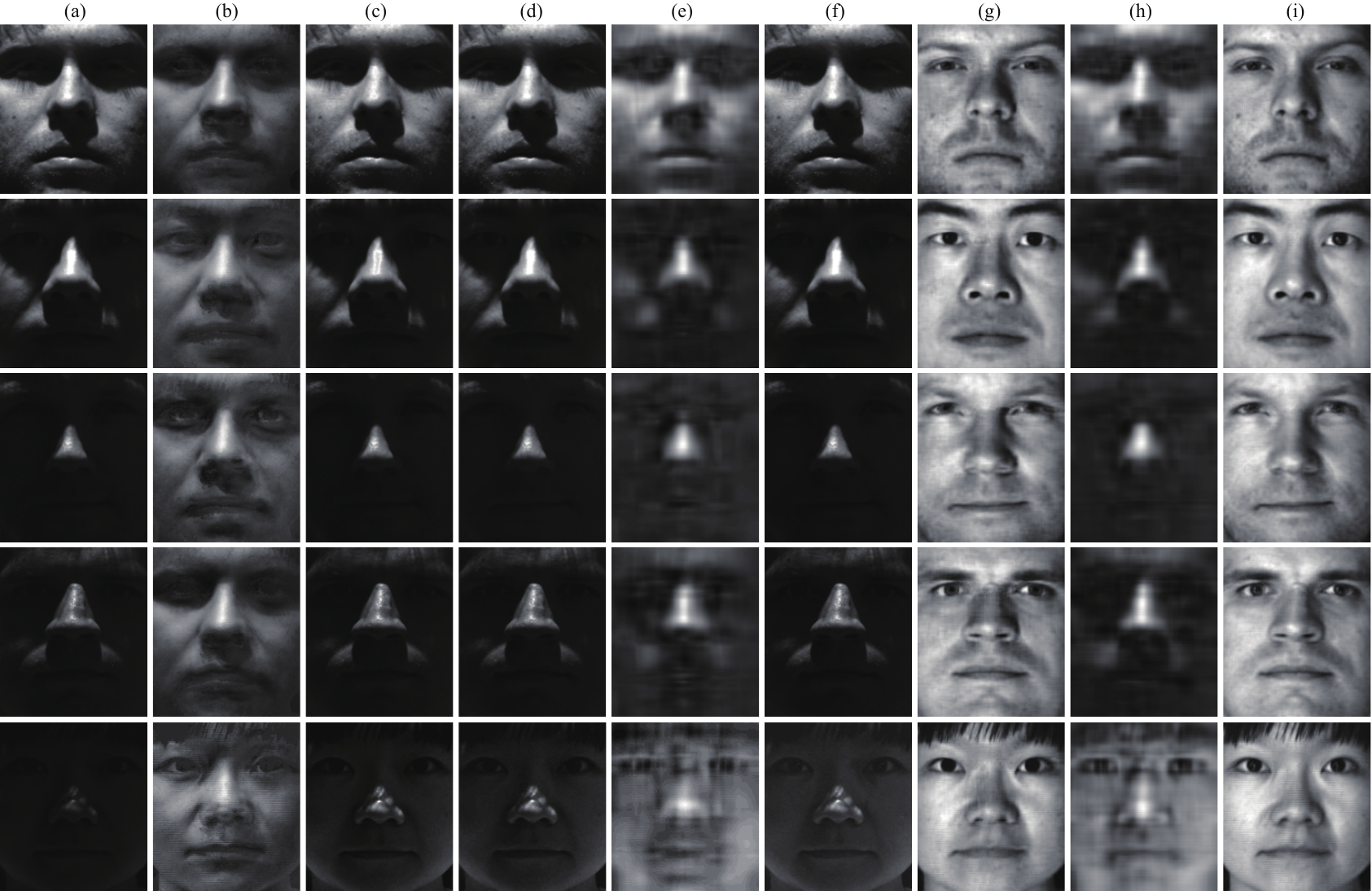}
   \caption{Face modeling results by different methods. (a)Original face images. (b)MoG LRMF. (c)HaLRTC. (d)LRTA. (e)PARAFAC. (f)MSI DL. (g)CWM LRTF. (h)MoG GWLRTF-CP. (i)MoG GWLRTF-Tucker.}
\label{fig_facemodeling}
\end{figure*}

\begin{figure*}[!t]
\centering
   \includegraphics[width=1.0\linewidth]{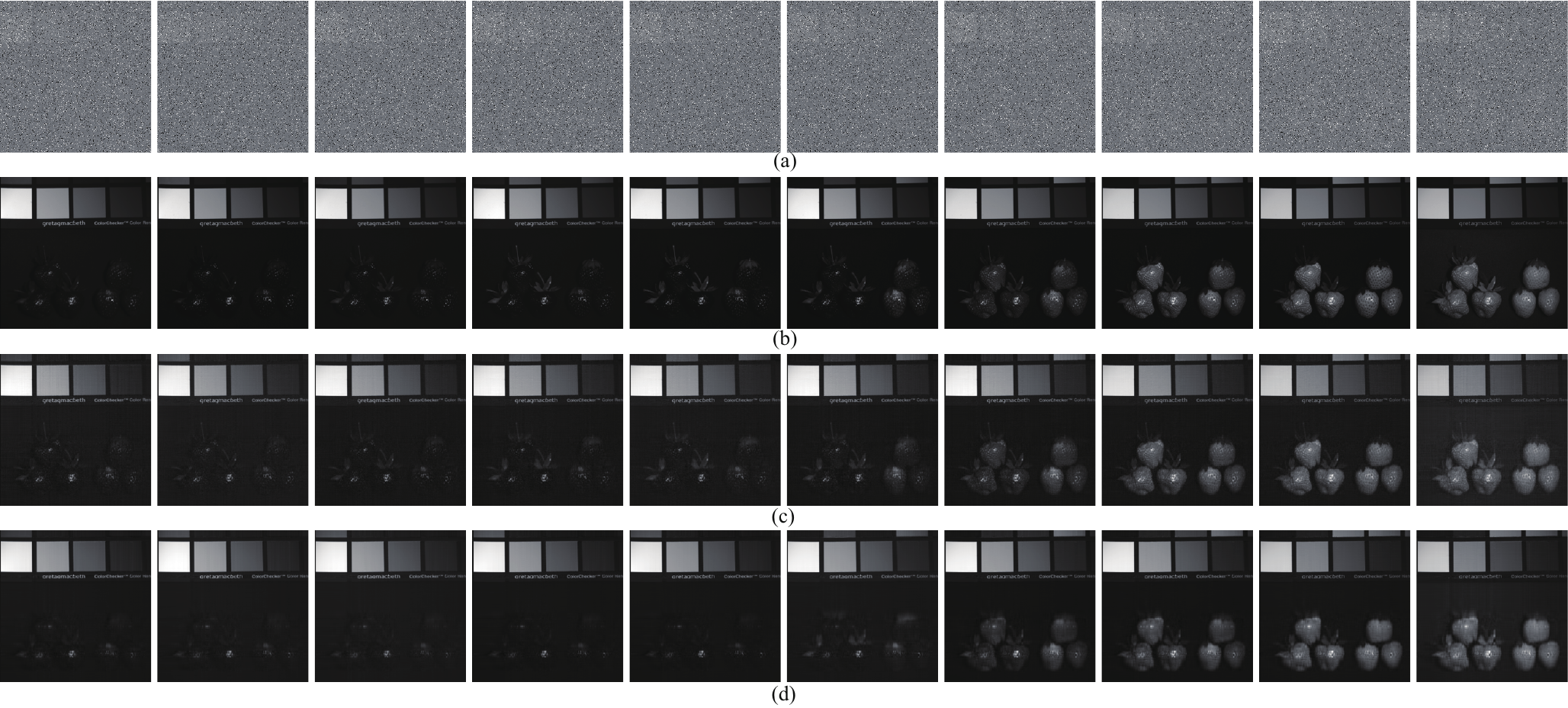}
   \caption{Ten randomly selected bands of strawberries. (a) Noisy bands. (b) Original bands. (c) Corresponding bands recovered by MoG GWLRTF-CP. (d) Corresponding bands recovered by MoG GWLRTF-Tucker.}
\label{fig_multispectralimages}
\end{figure*}
\begin{figure*}[!t]
\begin{center}
   \includegraphics[width=1.0\linewidth]{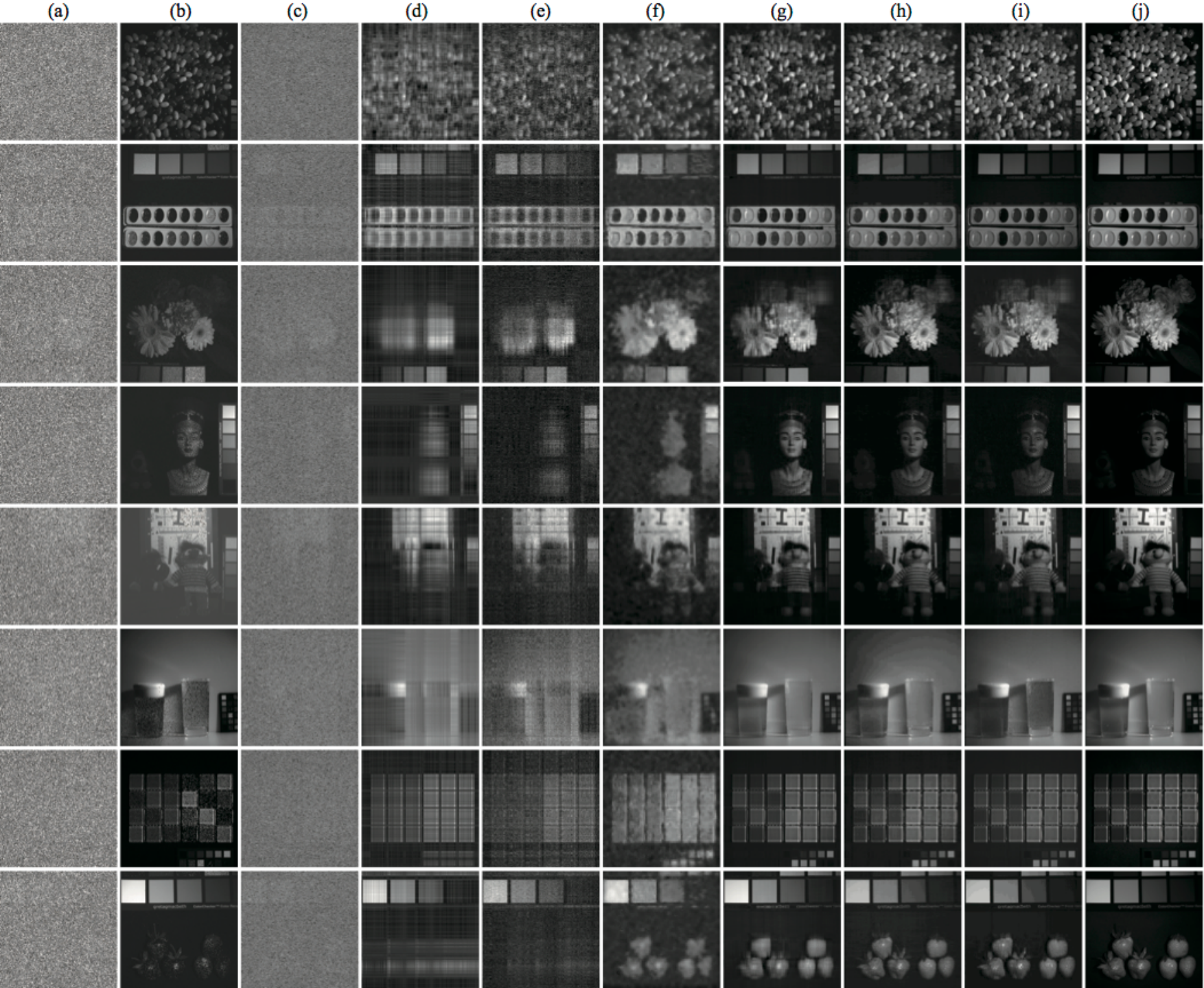}
\end{center}
   \caption{The 31st band of multispectral images. (a) Noisy band. (b)MoG LRMF. (c)HaLRTC. (d)LRTA. (e)PARAFAC. (f)MSI DL. (g)CWM LRTF.  (h)MoG GWLRTF-CP. (i)MoG GWLRTF-Tucker. (j) Original band.}
\label{fig_msirecovery}
\end{figure*}

\begin{table*}[!t]
\footnotesize
\caption{Multispectral image restoration performance of competing methods with mixture noise. In each experiments, the best performance is highlighted in bold and the second is underlined.}
\label{table4}
\centering
\begin{tabular}{c|ccccccccc}
\hline
         &       &MoG LRMF & HaLRTC&LRTA&PARAFAC& MSI DL&CWM LRTF  &MoG GWLRTF-CP&MoG GWLRTF-Tucker\\
  \hline
           &PSNR&   19.26  &  8.444  &16.51&15.84& 18.47&19.94&  \textbf{22.17} &  \underline{22.45} \\
Jelly     &RSE &  0.5169    & 1.773   &0.7003&0.7565& 0.5588&0.4720& \textbf{0.3652}& \underline{ 0.3535 }\\
  beans   &FSIM&  0.8101     &  0.5312  &0.7025&0.6487& 0.8213&0.8506& \textbf{0.8864}& \underline{0.9004 }  \\
\hline
         &PSNR&    18.90    & 8.725   &18.76&16.26& 19.20&21.75&\textbf{27.29} &  \underline{24.48}   \\
Paints     &RSE &  0.4022 &  1.336  &0.4209&0.5607& 0.3998&0.2983&\textbf{0.1576 } & \underline{0.2177}  \\
         &FSIM&    0.8543   &  0.4750  &0.7827&0.6367& 0.8182&0.9165&\textbf{0.9514}& \underline{0.9427}    \\
\hline
         &PSNR&  16.45    &  7.480  &18.40&16.54& 19.15&22.72&\textbf{ 26.17} &  \underline{ 24.24 }\\
Flowers     &RSE &  0.9073   &  2.522  &0.7173&0.8888& 0.6583&0.4364&\textbf{ 0.2933 } & \underline{0.3511} \\
         &FSIM&  0.8018    &  0.4172  &0.8073&0.5271& 0.8220&0.9126&\textbf{  0.9153}& \underline{0.9139 }   \\
\hline
         &PSNR&   15.32    &  7.109  &18.59&17.01& 19.43&23.20&\textbf{25.22 } &  \underline{23.58  } \\
Egyptian    &RSE &  1.563 &  3.847  &1.026&1.231& 0.9308&0.6032&\textbf{ 0.4779 } & \underline{0.5271  } \\
statue      &FSIM&    0.7834   & 0.3696   &0.8245&0.4243& 0.8142&0.9187&\textbf{ 0.9439}& \underline{ 0.9233}   \\
\hline
            &PSNR&   12.05         &  7.911  &18.29&16.23& 18.95&21.46& \textbf{ 23.95 }& \underline{ 23.33 }  \\
Chart    &RSE & 0.8633  &  1.388  &0.4201&0.5327& 0.3895&0.2919&\textbf{0.2191 }& \underline{ 0.2353 }  \\
         &FSIM&     0.8069   &  0.4534  &0.7729&0.5366& 0.8092&0.8901&\textbf{0.9221} & \underline{0.9402 }    \\
\hline
         &PSNR&  22.22     & 11.56   &19.92&17.14& 21.33&21.92&\textbf{ 25.07} &  20.98   \\
Beers     &RSE &  0.2389      & 0.8150   &0.3114&0.4287& 0.2646&0.2474&\textbf{0.1722} &  0.2756  \\
         &FSIM&   0.7659    &  0.4155 &0.7535&0.4423& 0.8214&\textbf{0.9430}&\underline{0.9200}&  0.8748  \\
\hline
         &PSNR&   19.02    &  8.380  &18.74&16.49& 18.89&\underline{21.66}&\textbf{26.66 }&  20.79  \\
Glass     &RSE & 0.6314 & 2.025   &0.6140&0.7961& 0.6038&\underline{0.4389}&\textbf{ 0.2467 } &  0.4541 \\
         &FSIM&  0.6990    &  0.4690  &0.7726&0.5738&0.7287&\underline{0.9075}&\textbf{0.9475 } &  0.7946 \\
\hline
         &PSNR&   17.06       &  7.762  &19.22&16.78& 19.54&18.99& \textbf{24.80} &  \underline{23.41 } \\
Strawberries &RSE &  0.7630 &  2.149  &0.5745&0.7607& 0.5534&0.5900&\textbf{0.3021}&\underline{ 0.3652}   \\
         &FSIM&   0.7906      & 0.3997   &0.8059&0.4932& 0.8128&\underline{0.9229}&\textbf{0.9283} &  0.9137  \\
\hline
\end{tabular}
\end{table*}

The synthetic tensor is generated as follows: firstly, matrices $\{U, V, T\}$ are drawn from a standard normal distribution, i.e., $\forall{i, j, k}$, the vectors $\bold{u}_{i}, \bold{v}_{j}, \bold{t}_{k}$ of the matrices $\{U, V, T\}$ comply with a standard normal distribution $\mathcal{N}(0,I_\mathbb{R})$; Secondly, construct the true tensor by $\mathcal{X}_{gt}=[\![U,V,T]\!]$, and set the size to $10\times10\times10$ and CP rank $r=5$. Then we conduct two synthetic experiments: i) for validating the predictive performance, we vary the true tensor missing entries rate ($20\%$, $40\%$, $60\%$) ; ii) for verifying the reconstruction performance, we randomly choose $20\%$ missing entries of the true tensor and further add certain type of noise to it as the following procedure: (1) Gaussian noise $\mathcal{N}(0,0.1)$; (2) Sparse noise: $20\%$ of the non-missing entries with the uniformly distribution over [-5,5]; (3) Mixture noise: $20\%$ of the non-missing elements with the uniformly distribution over [-5,5], and $20\%$ of the rest non-missing with Gaussian noise $\mathcal{N}(0,0.2)$ and the rest with $\mathcal{N}(0,0.01)$. The performance of each method is quantitatively assessed by the following measurements as used in~\cite{meng2013robust}:
 \begin{equation*}
  \begin{split}
  &E1={\|\mathcal{W}\odot({\mathcal{X}_{no}-\mathcal{X}_{rec}})\|}_{{L}_{1}},\\
  &E2={\|\mathcal{W}\odot({\mathcal{X}_{no}-\mathcal{X}_{rec}})\|}_{{L}_{2}},\\
   &E3={\|{\mathcal{X}_{gt}-\mathcal{X}_{rec}}\|}_{{L}_{1}},\\
   &E4={\|{\mathcal{X}_{gt}-\mathcal{X}_{rec}}\|}_{{L}_{2}},\\
  \end{split}
  \end{equation*}
 where ${\mathcal{X}_{no}}$ and ${\mathcal{X}_{rec}}$ are used to denote the noisy tensor and the recovered tensor, respectively. As mentioned in~\cite{meng2013robust}, $E1$ and $E2$ are the optimization objectives of existing methods, which assess how the reconstruction complies with the noisy input, but $E3$ and $E4$ are more meaningful for evaluating the correctness of the clean subspace recoveries.
 Therefore, we pay more attention to the quantitative indices of $E3$ and $E4$. In the tables, the first and second best performances are marked out with bold and underline, respectively.

 The performance of each method in the synthetic experiments are summarized in Table~\ref{table1} and Table~\ref{table2}, respectively. From Table~\ref{table1} we can see that, in the case of varying data missing rate, our methods always have a relative better predictive performance in all evaluation terms. When the data is only disturbed by a single distribution noise, i.e., Gaussian noise or sparse noise, other methods can also obtain a fairly well results. However, when the noise becomes complex, our methods still have a good reconstruction performance in this case, as shown in Table~\ref{table2}.

\subsection{Single RGB Image Reconstruction}

The benchmark colorful building facade image is used in this section to evaluate the performance of different methods in image reconstruction.
Note that the colorful image can be viewed as a 3-order tensor of size $493\times517\times3$.
Two groups of experiments are considered here.

Firstly, the facade image rescaled to [0,255] is randomly sampled with $20\%$ missing entries and then added with a relative small scale mixture noise: $20\%$ of the non-missing pixels with the uniformly distribution over [-35,35], $20\%$ of the rest non-missing pixels with Gaussian noise $\mathcal{N}(0,20)$ and the rest with another uniformly distribution $\mathcal{N}(0,10)$.

The visual effect of each methods are demonstrated in Figure~\ref{fig_facadelessnoise}. For better visual comparison, we have also provided a zoom-in version of a local region in Figure~\ref{fig_facadelessnoise}. From the results we can see that the MoG GWLRTF-Tucker method has a better reconstruction performance than the matrix based methods and the traditional tensor based methods in reconstructing the image details.

Besides the visual effect, quantitative assessments are also reported. 
Three quantitative image quality indices are adopted to evaluate the performance of each method: peak signal-to-noise ratio (PSNR), relative standard error (RSE) and feature similarity (FSIM) \cite{zhang2011fsim}.
Larger values of PSNR and FSIM and smaller values of RSE mean a better restoration results.

The quantitative results obtained by each method are given in Table~\ref{table3} (the upper row), and it shows that the MoG GWLRTF-Tucker method is superior to all the other methods except the MoG GWLRTF-CP method which performs nearly as well in this small scale mixture noise case.

Secondly, in order to further compare the reconstruction ability of each method, we add a larger mixture noise to the facade image. The image is first rescaled to [0,1] and then a larger mixture noise added as in the synthetic experiments: $20\%$ missing entries, $20\%$ of the non-missing pixels with the uniformly distribution over [-5,5], $20\%$ of the rest non-missing pixels with Gaussian noise $\mathcal{N}(0,0.2)$ and the rest with another uniformly distribution $\mathcal{N}(0,0.01)$.

Both the visual and the quantitative results are provided, as in Figure~\ref{fig_facade} and Table~\ref{table3} (the lower row). In Figure~\ref{fig_facade}, the zoom-in version of a local region for comparison is given. From the results, we can see that with the increasing of the mixture noise scale, the MoG GWLRTF-Tucker method performs much better than the MoG GWLRTF-CP method in reconstructing the image details. And they are both superior to the other methods which lose a lot of image structure information.

\subsection{Face Modeling}

In this section, we assess the effectiveness of the proposed methods in face modeling with different objects under different illuminations. It is different from the traditional methods in dealing with face modeling problem which always only focus on one kind of object under different illuminations. The dataset is the ensemble subset of the Extended Yale B database~\cite{georghiades2001few}, containing 45 faces of 5 objects and $9$ illuminations with size $192\times 168$. The original tensor is thus generated with size $192\times168\times 9\times 5$. All the competing methods used above are also compared here. Considering that some tensor methods are only designed to dealing with 3-order tensors and the matrix methods are originally designed to solve matrix data, in this case the original 4-order tensor is vectorized into the 3-order tensor with size $192\times168\times 45$ and matrix data with size $32256\times 45$ before processing, respectively.
The sampling images from this data subset are plotted in Figure~\ref{fig_oriface}.

Typical images and the face modeling results by all the competing methods are demonstrated in Figure~\ref{fig_facemodeling}. From Figure~\ref{fig_facemodeling}, we can see that the MoG GWLRTF-Tucker method outperforms the MoG GWLRTF-CP method a lot in this face modelling experiment and the CWM LRTF method achieves a fairly well result than the other methods which lose their efficiency in modelling different faces simultaneously. Note that the MoG GWLRTF-Tucker method is more sharper than the CWM LRTF method, especially in the cast shadows and saturations removing in the nose area of the faces. 

\subsection{Multispectral Image Recovery.}

The well-known Columbia Multispectral Image Database \cite{yasuma2010generalized}\footnote{\url{http://www1.cs.columbia.edu/CAVE/databases/multispectral}} contains 32 scenes of a wide range of real world objects with image size $512\times512$ and 31 spectral bands. Here we use 8 of them (Jelly beans, Paints, Flowers, Egyptian statue, Chart and stuffed toy, Beers, Glass tiles, Strawberries) to test the efficiency of our methods. In this experiment, each of these MSIs is resized to half and rescaled to [0,1]. Then the same large mixture noise are added to the images as in the single RGB image reconstruction experiment.

For better visual demonstration of the multispectral image restoration results, we randomly choose ten selected bands of the strawberries for shown. The noisy bands, original bands and their corresponding bands recovered by the MoG GWLRTF-CP method and the MoG GWLRTF-Tucker method are shown in Figure~\ref{fig_multispectralimages}.
Meanwhile, the 31st band of these multispectral images are selected to show our restoration results compared with other competing methods in Figure~\ref{fig_msirecovery}. The quantitative indices of PSNR, RSE and FSIM are also used here to further evaluate the effectiveness of each method and the quantitative results are given in Table~\ref{table4}.

From Figure~\ref{fig_msirecovery} and Table~\ref{table4}, we can see that the MoG GWLRTF-CP method outperforms the MoG GWLRTF-Tucker method in recovering the multispectral images and both of them obtain a better result than the other competing methods. It is worth to mention that while the matrix based method MoG LRMF obtain a roughly good results on certain multispectral images (eg. the Jelly beans and the Paints), all the 31 bands they recovered from the corresponding multispectral image are all the same which indeed lose the specific information of each band.


\subsection{Real Hyperspectral Image Restoration}

In this section, we further apply the proposed methods, i.e., MoG GWLRTF-CP and MoG GWLRTF-Tucker, to the real hyperspectral image restoration application to test their efficiency compared with other methods. Here we use a HYDICE urban image\footnote{\url{http://www.tec.army.mil/hypercube}} for demonstration and it contains 210 bands with size $307\times307$. However, several bands of them are seriously polluted by the atmosphere and water absorption. Generally, most existing methods choose to discard these seriously polluted bands before applying their methods to restoring the rest of these bands as in~\cite{zhao2015novel}. Different from them, we directly apply our methods to all these bands and try to restore all of them. 

\begin{figure*}[!t]
\begin{center}
   \includegraphics[width=1.0\linewidth]{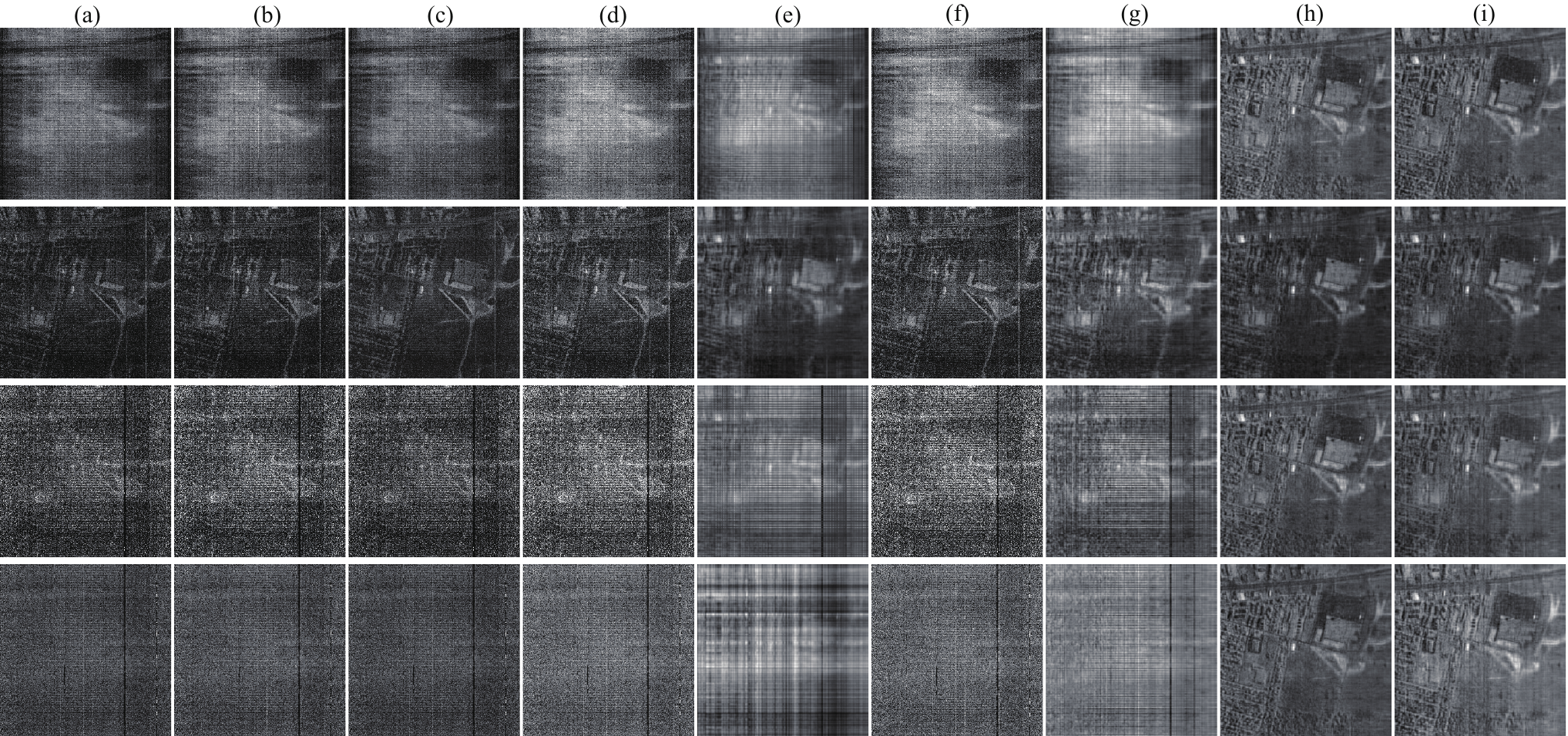}
\end{center}
   \caption{Real hyperspectral image restoration. (a) Original polluted bands. (b)MoG LRMF. (c)HaLRTC. (d)LRTA. (e)PARAFAC. (f)MSI DL. (g)CWM LRTF.  (h)MoG GWLRTF-CP. (i)MoG GWLRTF-Tucker. }
\label{fig_realimage}
\end{figure*}

In Figure~\ref{fig_realimage}, the first column demonstrates four seriously polluted bands in HYDICE urban image. And the corresponding restoration results obtained by MoG GWLRTF-CP, MoG GWLRTF-Tucker and other competing methods are given in the following columns. The results indicate that MoG GWLRTF-CP is a little better than MoG GWLRTF in the restoration of these polluted real hyperspectral bands, while the other competing methods are basically ineffective.

\section{Conclusion and Discussion}

In this paper, as an extension of our last work \cite{chen2016robust}, we propose a generalized weighted low-rank tensor factorization method integrated with MoG (MoG GWLRTF). It contains the MoG GWLRTF-CP method and the MoG GWLRTF-Tucker method, and both of them can more sharperly estimate the subspaces from high-dimensional data which may be polluted by noise with complex distribution. And the corresponding algorithms designed under the EM framework are proposed to solve this two subproblems. Extensive experiments are conducted to validate the efficiency of this two methods and some other existing matrix based methods and tensor based methods are also compared.

The results show that both the MoG GWLRTF-CP method and the MoG GWLRTF-Tucker method are not only capable of better preserving the image structure information but also performing better when the data are disturbed by a large percentage of complex noise compared with other competing methods. Meanwhile, we find that the MoG GWLRTF-CP method and the MoG GWLRTF-Tucker method have show their advantages in dealing with different kinds of applications. To our knowledge, the MoG GWLRTF-Tucker method is based on the Tucker factorization and its core tensor is better at controlling the interaction of each factor matrices while the MoG GWLRTF-CP method is based on the number of rank-1 tensors which is good for sparse compressing problems.

Inspired by this phenomenon, we will further investigate the difference between the MoG GWLRTF-CP method and the MoG GWLRTF-Tucker method by applying them to more real applications and integrated them with the markov random field (MRF) to study their performance in dealing with high-order video data.

\section*{Acknowledgment}

This work was supported by the National Natural Science Foundation of China (Grant No. 61303168, 61333019, 11501440, 61373114).

\ifCLASSOPTIONcaptionsoff
  \newpage
\fi



\bibliographystyle{IEEEtran}
\bibliography{IEEEabrv,mybibfile}
\end{document}